\documentclass[runningheads]{llncs}
\usepackage[T1]{fontenc}
\usepackage{graphicx}
\usepackage{subcaption} 
\usepackage{float} 
\usepackage{pdflscape} 
\usepackage{adjustbox}

\usepackage{booktabs}
\usepackage{graphicx}
\usepackage{tabularx}
\usepackage{multirow}
\usepackage{amsmath}

\usepackage{lineno}
\usepackage{float}
\title{Pustak AI: Curriculum-Aligned and Interactive Textbooks Using Large Language Models}
\titlerunning{PustakAI}

\author{
Shivam Sharma \and Riya Naik \and
Tejas Gawas \and
Heramb Patil \and 
Kunal Korgaonkar
}

\institute{
 Department of Computer Science and Information Systems, \\ BITS Pilani K K Birla Goa Campus, Goa, India \\
\email{Email: kunalk@goa.bits-pilani.ac.in}
}

\begin{document}
\maketitle
\begin{abstract}
Large Language Models (LLMs) have demonstrated remarkable capabilities in understanding and generating human-like content. This has revolutionized various sectors such as healthcare, software development, and education. In education, LLMs offer potential for personalized and interactive learning experiences, especially in regions with limited teaching resources. However, adapting these models effectively to curriculum-specific content, such as the National Council of Educational Research and Training (NCERT) syllabus in India, presents unique challenges in terms of accuracy, alignment, and pedagogical relevance.
In this paper, we present the framework "PustakAI"\footnote{Pustak means `book' in many Indian languages.} for the design and evaluation of a novel question-answering dataset "NCERT-QA" aligned with the NCERT curriculum for English and Science subjects of grades 6 to 8. We classify the curated QA pairs as Factoid, Inferential, and Others (evaluative and reasoning). We evaluate the dataset with various prompting techniques, such as meta-prompt, few-shot, and CoT-style prompting, using diverse evaluation metrics to understand which approach aligns more efficiently with the structure and demands of the curriculum. 
Along with the usability of the dataset, we analyze the strengths and limitations of current open-source LLMs (Gemma3:1b, Llama3.2:3b, and Nemotron-mini:4b) and high-end LLMs (Llama-4-Scout-17B and Deepseek-r1-70B) as AI-based learning tools in formal education systems.

\keywords{Large Language Model  \and QA Systems \and Educational AI.}

\end{abstract}

\section{Introduction}

The idea of a machine that could answer questions, write essays, or translate natural language like humans seemed fiction. But with the advent of Large Language Models (LLMs), that vision has become reality. LLMs built on a vast amount of data learn using transformer architectures. As they evolved, LLMs have been used to assist in various tasks such as composing emails, writing code, scientific discovery, and tutoring. It is the last category that is of interest to us. 
Trained on a diverse range of subjects and answering styles, LLMs can assist students by answering questions, explaining complex concepts, and even offering feedback on their submissions. On the other hand, for educators, it can help create lesson plans, automate learning tasks such as creating balanced questionnaires, and also explain concepts with additional research. LLMs have the potential to benefit education by extending learning beyond standard teaching-learning and help bridge the educational gap. To facilitate this, increasing integration of language models into educational contexts has prompted a wave of research exploring their capabilities, limitations, and impact.  

Researchers and developers are utilizing LLMs to create interactive learning platforms that can adapt to student needs. These models have been used to generate practice questions, summarize complex topics, provide coding help, and translate languages, thereby supporting diverse learners across disciplines. However, the challenge of efficiently training LLMs for educational purposes remains. This is largely due to the quality of training data and the inference methods employed. Recent progress in dataset development has focused on general educational content, but for practical use in institutional settings, the data must be tailored to specific curriculam. Therefore, it is equally important to improve LLMs for educational question answering by introducing more effective inference methods and creating high-quality, curriculum-aligned datasets. This will enable the fine-tuning of both LLMs and traditional language models, leading to more accurate and contextually relevant educational responses.

Our framework PustakAI aims to contribute as follows:
\begin{itemize}
    \item Present the \textbf{NCERT-QA} dataset to implement a curriculum-aligned Q\&A system. This study validates the NCERT-QA dataset as a foundation for building a curriculum-aligned Q\&A system. We develop a QA dataset derived directly from curriculum content and conduct a comprehensive evaluation of its effectiveness through various inference prompting techniques.
    \item Demonstrate the unique challenge posed by our curriculum-specific dataset by baselining it against a general-domain benchmark such as SQuAD.
    \item Perform an in-depth analysis to identify the optimal models and prompting strategies for the NCERT-QA task, including a comparison between subjects
    \item Analyze the practical trade-offs between performance and efficiency to make a case for cost-effective deployment in real-world school settings.
\end{itemize}
The rest of the paper is organized as follows. Section \ref{sec:rel} summarizes existing educational datasets, Sections \ref{sec:datacu} and \ref{sec:eda} elaborate the dataset curation steps and dataset analysis, Section \ref{sec:method} describes the implementation of the pipeline to evaluate the dataset and inference strategies, and Section \ref{sec:exp} summarizes the experimental evaluation and results. Finally, Section \ref{sec:conc} summarizes the conclusions drawn. 

\section{Background and Related Work}
\label{sec:rel}
The application of LLMs to question answering has evolved from general benchmarks like SQuAD \cite{lai2017race} to high-stakes domains like education, where the risks of "hallucination" demand high factual accuracy and pedagogical alignment. This has driven research in two key areas: the creation of curriculum-aligned datasets and the development of methodologies like advanced prompting to ensure model outputs are faithful to reliable sources.

Early educational datasets focused on broad reasoning, such as ARC for science \cite{clark2018arc} and FairytaleQA for narrative comprehension \cite{xu2022fantastic}. A move toward direct curriculum alignment was marked by datasets like RACE, sourced from student examinations \cite{lai2017race}. This trend has intensified with resources tightly coupled to specific textbook content, such as CK12-QA for science \cite{alawwad2025evaluating} and PeerQA for scientific reviews \cite{baumgartner2025peerqa}. Concurrently, the focus on evaluation has sharpened, with benchmarks like SyllabusQA introducing fact-checking metrics \cite{fernandez2024syllabusqa} and TruthfulQA testing models against common falsehoods \cite{lin2021truthfulqa}. Despite this progress, a significant gap remains for a large-scale dataset aligned with a major non-Western curriculum like India's NCERT, a gap our work aims to fill. A comparative analysis of these datasets is provided in (Table \ref{tab:dataset_comparison}).

Aligning LLMs to be pedagogically sound is a key challenge, as general-purpose models are not inherently suited for the classroom. Frameworks like COGENT demonstrate how to generate grade-appropriate content by providing structured guidance on learning objectives and readability \cite{liu2025cogent}. This has also prompted architectural debates, contrasting large unified models with more efficient Mixture-of-Experts (MoE) architectures tailored to specific curricula \cite{razafinirina2024pedagogical}. Our evaluation of a wide spectrum of models contributes directly to this investigation, providing empirical data on the performance-cost trade-offs for deploying practical AI tools in school systems.

In education, ensuring faithfulness (grounding answers in trusted sources) is non-negotiable. The standard approach combines Retrieval-Augmented Generation (RAG) for its architecture \cite{lewis2020retrieval} with advanced prompting to control the model's reasoning process. While Chain-of-Thought (CoT) was an early breakthrough for eliciting reasoning \cite{kojima2022large}, the efficacy of popular frameworks like ReAct has been challenged. A recent critical evaluation found that ReAct's performance gains stem from exemplar similarity rather than genuine reasoning, revealing a failure to generalize \cite{bhambri2025react}. This critique highlights the need for robust alternatives like meta-prompting, which uses high-level, structural guidance to enforce a faithful reasoning process \cite{zhang2024meta}. Our work directly investigates if this structural approach is more effective than the content-based guidance of few-shot or CoT prompts in a curriculum-aligned context.

\begin{table}
\centering
\caption{Comparative Analysis of Key Question-Answering Datasets.}
\label{tab:dataset_comparison}
\resizebox{\textwidth}{!}{%
\begin{tabular}{@{}l|p{2cm}|p{2cm}|p{2cm}|l|p{2.5cm}|p{5cm}@{}}
\toprule
\textbf{Dataset} & \textbf{Focus} & \textbf{Target Age/Grade} & \textbf{Subjects} & \textbf{Lang.} & \textbf{Curriculum Alignment} & \textbf{Key Features / Relevance} \\
\midrule
SQuAD & Extractive QA & General Adult & General (Wikipedia) & English & None & Foundational extractive QA benchmark. \\ \hline
RACE & Reading Comprehension & Grades 7-12 (Ages 12-18) & English & English & High (Chinese Examinations) & Precedent for curriculum-aligned QA from exams. \\ \hline
ARC & Science Reasoning & Grades 3-9 & Science & English & Loose (Grade-level science) & Benchmark for complex science reasoning. \\ \hline
SciQ & Science QA & General & Science (Physics, Chem, Bio) & English & Loose (General science topics) & Provides supporting evidence text with questions. \\ \hline
TruthfulQA & Factual Faithfulness & General Adult & General (38 categories) & English & N/A & Measures model's ability to avoid common falsehoods. \\ \hline
FairytaleQA & Narrative Comprehension & Grades K-8 & Reading/ Stories & English & None & Expert-generated questions for younger students. \\ \hline
CK12-QA & Multimodal Textbook QA & Middle School & Science & English & High (CK-12 Textbooks) & Direct parallel for RAG on science textbooks. \\ \hline
SyllabusQA & Course Logistics QA & University & General (36 majors) & English & High (University Syllabi) & Introduces Fact-QA metric for factual accuracy. \\ \hline
PeerQA & Scientific Document QA & Graduate+ & STEM/NLP & English & N/A (Scientific Papers) & Expert-generated questions from authentic sources. \\ 
\midrule
\textbf{NCERT-QA} & \textbf{Curriculum-aligned QA} & \textbf{Grades 6-8} & \textbf{Science, English} & \textbf{English} & \textbf{High (Indian NCERT)} & \textbf{Addresses the gap for a major non-Western curriculum.} \\ 

\bottomrule
\end{tabular}%
}
\end{table}

\section{Dataset Curation}
\label{sec:datacu}

\begin{figure}[htbp]
\centering
\begin{minipage}{0.48\textwidth}
\centering
\includegraphics[width=\linewidth]{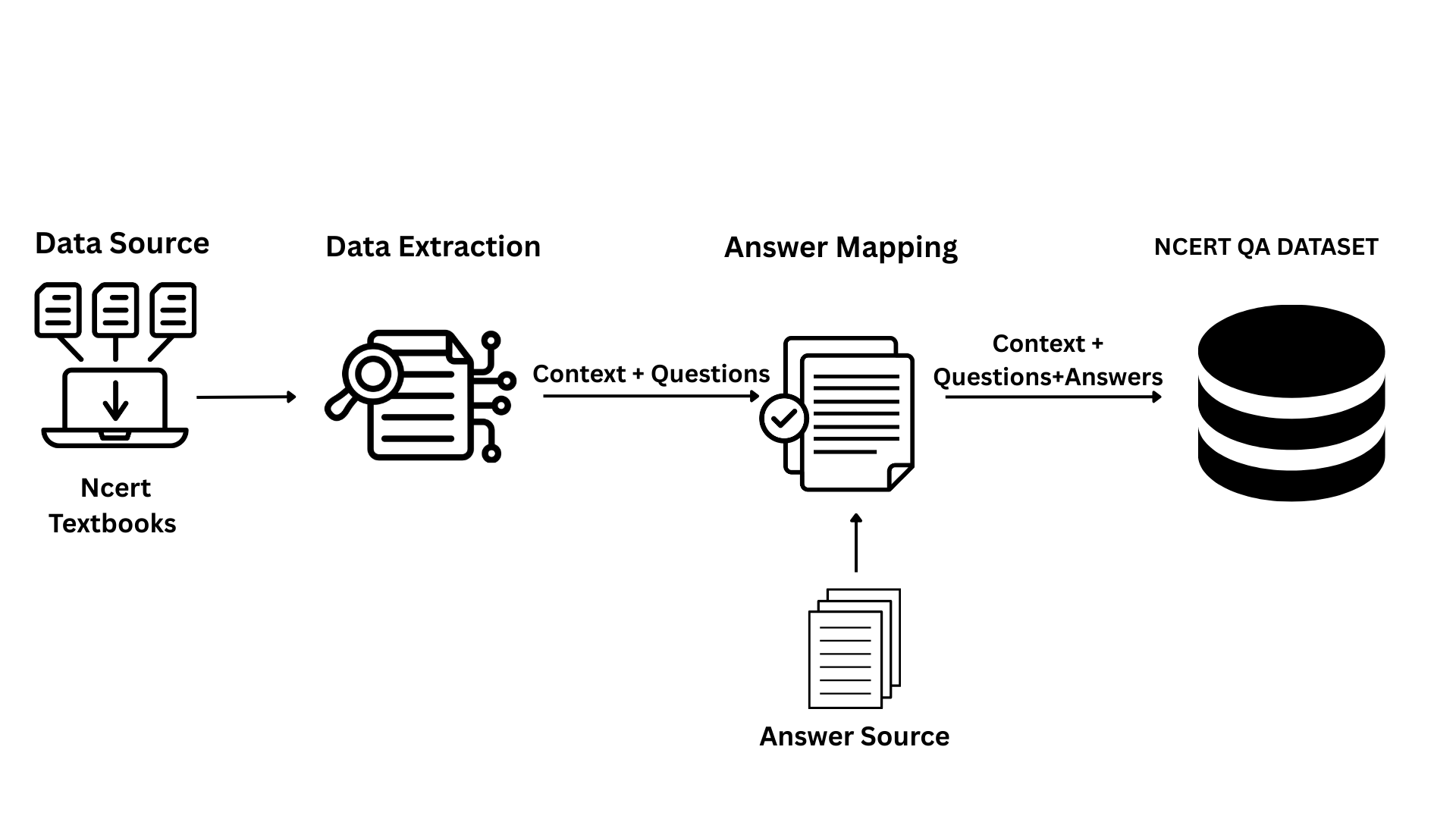}
\textbf{(a)}

\label{fig:data}
\end{minipage}%
\hfill
\begin{minipage}{0.48\textwidth}
\centering
\includegraphics[width=\linewidth]{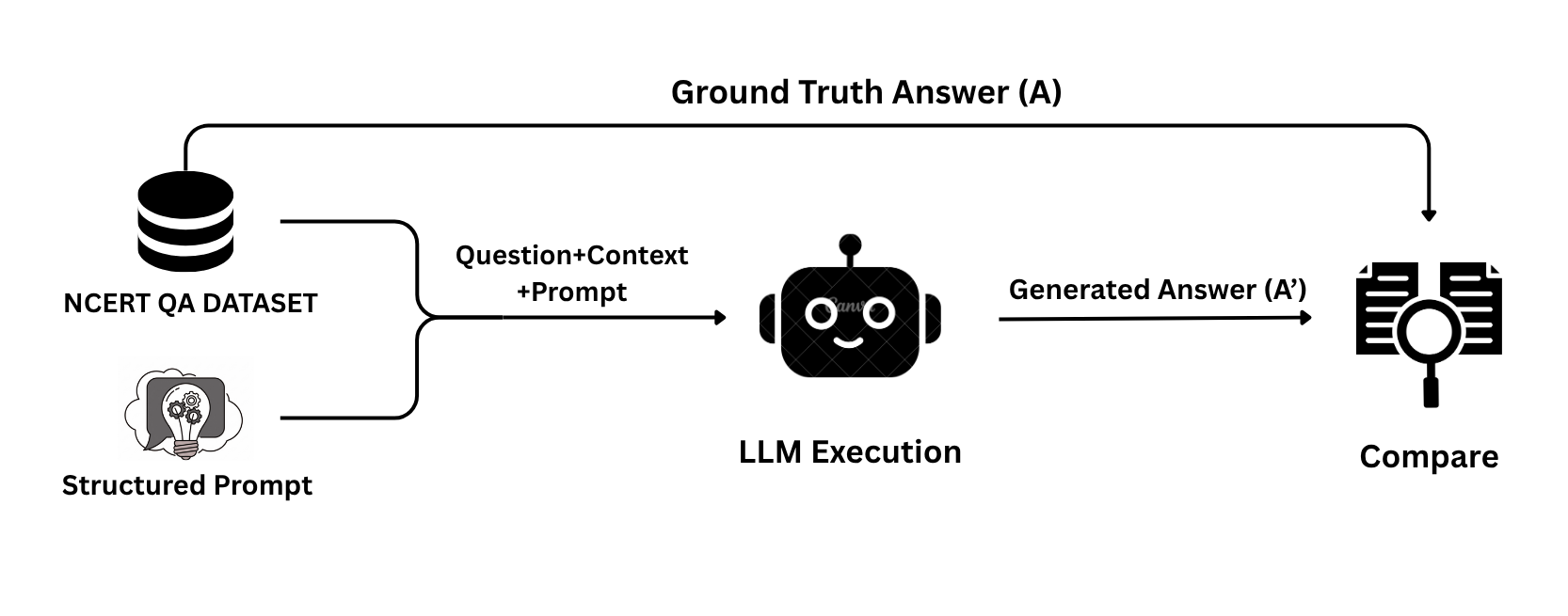}
\textbf{(b)} 

\label{fig:llmpipeline}
\end{minipage}
\caption{(a) NCERT-QA dataset curation process. NCERT textbooks are parsed to extract chapters as context and the respective questions. Answers are retrieved from various authentic public online sources and aligned based on chapter and question indices. These answers are used as ground truth. The resulting QA dataset is structured as a collection of context-question-answer tuples. (b)  LLM prompting and evaluation pipeline. LLM is presented with chapter and its corresponding question by employing a variety of prompting strategies. Model then generates a response to the question which is compared against ground truth using various evaluation matrices.}
\label{fig:combined}
\end{figure}

Our NCERT-QA data curation process includes three major steps: collection of NCERT text documents, data extraction, and answer mapping. Each of the steps is detailed below and is visualized in Fig. \ref{fig:combined}(a).
Our objective is to collect high-quality, authentic data. To achieve this, we gathered 35 documents (chapters) from the English curriculum and 48 documents from the Science curriculum for classes 6 to 8. Each document’s textual content was meticulously extracted from PDFs, with all images, captions, and tables removed to ensure the purity of the text. Our overall refined corpus comprised a total of 83 documents.

\subsection{Data Extraction}
In each chapter, we systematically extracted the chapter text as context and the questions provided in the exercises of the respective chapter. Specifically, a total of 451 questions were extracted for English documents, while 288 questions were obtained from the science documents, bringing the dataset to a total of 739 question-answer pairs. These extracted questions were subsequently categorized into three distinct types: Factoid, Inferential, and a third category denoted as Others.
The \textbf{Factoid} category consists of questions whose answers can be directly extracted from the passage, identified as specific spans of text. 
The \textbf{Inferential} category includes questions that necessitate logical reasoning and inferential thinking to develop a comprehensive response based on the passage content. 
The \textbf{Other} category is distinct from the first two; it encompasses questions that are boolean in nature or require answers extracted from multiple paragraphs, among other traits. Examples for each category are illustrated in Table \ref{tab:question_categories}(a)

\begin{table}[htbp]
\centering
\tiny
\caption{(a) Examples of different question categories with corresponding answers; (b) Distribution of Question Types in the NCERT-QA Dataset.}
\begin{minipage}[t]{0.48\textwidth}
\centering
\label{tab:question_categories}
\begin{tabular}{l|p{2.0cm}|p{2.0cm}}
\toprule
\textbf{Category} & \textbf{Question} & \textbf{Answer} \\
\midrule
Factoid & What did Patrick think his cat was playing with? What was it really? & Patrick thought his cat was playing with a doll, but it was actually a tiny man. \\
\hline
Inferential & Why did the little man grant Patrick a wish? & Because Patrick saved him from the cat and the elf wanted to return the favor. \\
\hline
Others & In what way did the shopkeeper make a fool of Rasheed? & The shopkeeper pretended Rasheed could win prizes, but tricked him with cheap goods. \\
\bottomrule
\end{tabular}
\end{minipage}%
\hfill
\begin{minipage}[t]{0.48\textwidth}
\centering
\label{tab:qa_distribution}
\begin{tabular}{@{}lcc@{}}
\toprule
\textbf{Question Category} & \textbf{Count} & \textbf{Percentage} \\
\midrule
Factoid & 405 & 55\% \\
Inferential & 258 & 35\% \\
Other & 76 & 10\% \\
\midrule
\textbf{Total} & \textbf{739} & \textbf{100\%} \\
\bottomrule
\end{tabular}
\end{minipage}
\end{table}

\subsection{Answer Mapping}
In the final phase of dataset curation, we execute the process of correlating answers to the questions extracted from the exercises. We utilize the solutions to each specified question sourced from authentic public online sources. The mapping of questions is achieved by utilizing the chapter index and question number as reference points. Subsequently, we manually assess the accuracy and appropriateness of answer mappings to ensure their correctness.

\section{Exploratory Dataset Analysis}
\label{sec:eda}

In this section, we conduct an in-depth analysis of the NCERT-QA dataset to better understand its defining characteristics and underlying structure. Our primary focus lies in examining the diversity of the data and the formulation of question-answer pairs. This analysis is crucial for evaluating the dataset’s suitability for building robust educational QA systems and for highlighting how its features can be used to test different model capabilities.

\subsection{Data Diversity}
\begin{figure}
    \centering
    \includegraphics[width=1\textwidth]{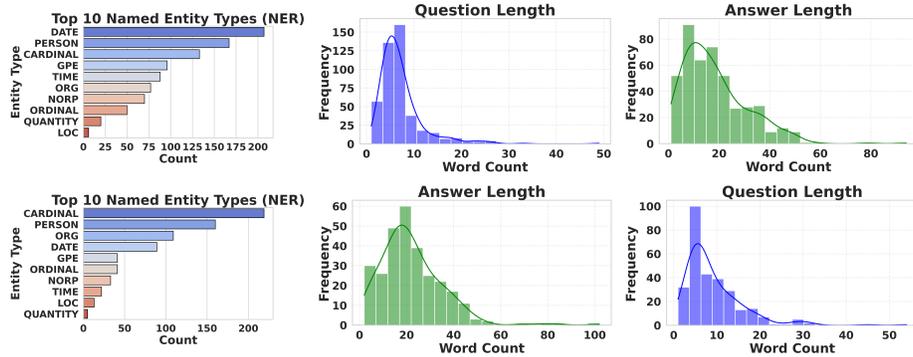}
    \caption{Subject-wise analysis of the NCERT-QA dataset showing question\&answer length distributions and top named entities, highlighting differences in complexity and focus between English and Science.}
    \label{fig:eda}
\end{figure}

The NCERT-QA dataset is intentionally diverse, covering two distinct subjects English and Science across three consecutive grade levels (6, 7, and 8). This subject diversity introduces a variety of text complexities and styles. The English texts are primarily narrative and literary, requiring comprehension of plot, character, and thematic elements. In contrast, the Science texts are descriptive and explanatory, demanding an understanding of concepts, processes, and factual information. This is evident from the Q\&A length distribution shown in Fig \ref{fig:eda}. The data indicate that Science subject documents tend to feature more elaborate questions and answers compared to those in English. Most English questions fall within the 5–10 word range, while Science questions typically range from 10–20 words, reflecting their more conceptually driven nature. Similarly, the answer length distribution highlights the explanatory style of Science responses, with answer lengths commonly falling between 10-40 words. This dichotomy provides a comprehensive testbed for evaluating an LLM's adaptability to different linguistic domains and reasoning types within a single, coherent educational framework. 
Although the two subjects differ in the structure and flow of their texts, they also exhibit notable similarities. As shown in Fig \ref{fig:eda}, both English and Science texts frequently reference entities such as PERSON, CARDINAL, DATE, and ORG. This suggests that while English and Science differ in textual structure and cognitive demands, they converge in their reference to certain key real-world entities which may aid in the development of cross-domain entity recognition and information extraction capabilities in language models.

\subsection{Question-Answer Formulation}
The formulation of the question-answer pairs is designed to mirror real-world educational exercises and assess a range of cognitive skills. As introduced in Section \ref{sec:datacu}, questions are categorized as Factoid, Inferential, and Other. A quantitative breakdown of the 739 questions in our dataset reveals the distribution shown in Table \ref{tab:qa_distribution}(b). The prevalence of Factoid questions (55\%) ensures a solid baseline for testing a model's core reading comprehension and information retrieval abilities. A significant portion of Inferential questions (35\%) is critical for evaluating deeper reasoning, requiring models to connect ideas and make logical deductions that are not explicitly stated in the text. The Other category (10\%) includes more complex questions that often require synthesizing information from multiple paragraphs, providing a challenge for advanced reasoning. This balanced distribution ensures that our evaluation rigorously tests models on a spectrum of tasks, from simple lookup to complex synthesis, which is essential for a versatile educational assistant.

\section{Methodology}
\label{sec:method}
In this study, we establish a pipeline founded on LLMs to exhibit the utilization of the NCERT-QA dataset. A comparative analysis is conducted on the performance of various LLMs, which include (gemma3:1b, llama3.2:3b, and nemotron-mini:4b) and high-end LLMs (Llama-4-Scout-17B-16E-Instruct and deepseek-r1-distill-llama-70b) with the objective of evaluating the enhancement in performance facilitated by this dataset, as shown in the Fig \ref{fig:combined}(b).

Each input to the LLM consists of a question paired with its corresponding context. We wrap this combination into a prompt to guide the model through a reasoning process. Prompt is stitched in a way that the model first determines the type of question—whether it is factoid, inferential, or others. Based on the identified question type, the model is then expected to follow appropriate reasoning steps using the given context to arrive at an accurate answer.

After the LLM generates an answer, we evaluate the answer against the ground truth answers provided in the NCERT-QA dataset. We employ a range of evaluation metrics to assess the quality of the responses. These metrics go beyond simple word overlap measures like ROUGE-L\cite{lin-2004-rouge} to contextual relevance using BERTScore-f1\cite{zhang2019bertscore}. 

\noindent
We extend the comparison by evaluating the answer quality in terms of faithfulness and semantic relevance. We calculate semantic relevance as cosine similarity between sentence embeddings using SentenceTransformer models\cite{reimers2019sentence}:
\[
\text{Faithfulness} =
\frac{|tokens_{\text{answer}} \cap tokens_{\text{context}}|}{|tokens_{\text{answer}}|}
\quad \quad
\text{SemanticSim} =
\frac{e_{\text{ref}} \cdot e_{\text{gen}}}{\lVert e_{\text{ref}}\rVert \, \lVert e_{\text{gen}}\rVert}
\]

 \textit{SemanticSim} (SS) measures how well the generated answer aligns with the intended meaning of the reference answer, regardless of exact word matches. Unlike traditional lexical metrics such as ROUGE, semantic similarity evaluates \textit{conceptual coherence} between responses using cosine similarity between sentence embeddings.

\textit{Faithfulness} (FF) measures the extent to which generated answers remain grounded in the provided source context, critical for educational applications where accuracy is paramount. This multilayered evaluation ensures a robust assessment of the LLM's ability to understand, reason, and respond accurately to NCERT-QA.

\subsection{Prompt Strategies}
We implement the framework detailed in \ref{sec:method} using diverse state-of-the-art prompting strategies \cite{wei2022chain,dang2022prompt}. This implementation is designed to evaluate the underlying inference process utilizing our dataset.

\noindent
The prompting strategies used are as follows:
\begin{enumerate}
    \item Shot-based: We provide one example of each question category for the LLM to understand the answering pattern before asking the model to perform answer retrieval. This helps guide the model by showing the format and logic needed to come up with a response, without overloading it with lots of interleaved instructions. We extend this prompt to 3 and 5 shots by increasing the number of examples such that LLM can learn the diverse spectrum of each category.
    \item CoT: In this strategy, we provide a method that encourages the model to reason step by step rather than jumping straight to the answer. We utilize this method to solve complex problems that require logical reasoning, intermediate steps, or multi-hop reasoning within the context.
    \item Meta Prompt (MP): Our approach involves employing the large language model, Claude-Sonnet, to construct a prompt or template that is used in our subsequent NCERT question and answer task. The goal of incorporating a meta\_prompt is to evaluate whether Instructions by a meta-language model can enhance generating responses from the answering LLM, as opposed to relying solely on human-devised instructions.
    \item Meta One-shot (MP-1S): In this work, we utilize a dual approach by integrating a meta prompt with one illustrative example of question types. To construct the prompt, we begin by drawing upon instructions sourced from a meta-level language model and subsequently embed an instance for each category of question within the prompt as examples.
\end{enumerate}

\section{Results and Analysis}
\label{sec:exp}
Our experimental results are presented in three parts. We first establish the unique value of the NCERT-QA dataset, then analyze model and prompt performance on our task, and finally discuss the practical implications for deployment.

\noindent
\textbf{Part 1: The Unique Challenge of Curriculum-Aligned QA:}
To demonstrate that answering curriculum-specific questions is a distinct challenge that general-purpose models cannot solve from pre-trained knowledge alone, we conducted a baseline experiment. We evaluated Llama4-Scout-17B on the well-known SQuAD dataset and on our NCERT-QA dataset without providing the textbook context. This setup forces the model to rely solely on its internal knowledge.

The results in Table \ref{tab:squad_vs_ncert} are unequivocal. The model performs exceptionally well on SQuAD, a general-knowledge benchmark and possibly used for LLM training, but its performance collapses on NCERT-QA questions when deprived of the textbook context. The F1 score plummets, indicating an inability to generate precise answers. This is because NCERT questions are deeply tied to the specific phrasing, narratives, and vocabulary of the curriculum, which is not adequately represented in the model's general training data. This experiment confirms our core hypothesis: a specialized dataset and prompting with the right context are not just helpful but essential for building a reliable educational assistant.

\begin{table}[htbp]
\centering
\scriptsize
\caption{Baseline performance of Llama-4-Scout-17B on SQuAD vs. NCERT-QA (no context), showing the performance drop and the need for a curriculum-specific, context-aware approach.}
\label{tab:squad_vs_ncert}
\begin{tabular}{l|c|c|c|c}
\toprule
\textbf{Metric} & \textbf{SQuAD} & \textbf{NCERT QA (Eng)} & \textbf{NCERT QA (Sci)} & \textbf{NCERT-QA (Overall)} \\
\midrule
F1      & \textbf{0.894} & 0.32 & 0.47 & 0.395 \\
Semantic Sim. & \textbf{0.920} & 0.78 & 0.88 & 0.830 \\
\bottomrule
\end{tabular}
\end{table}

We further assessed the models using each prompt type with and without incorporating contextual information on NCERT-QA. As shown in Tables \ref{tab:nocon} and \ref{tab:bestperf}, there is a noticeable improvement in metric scores when context is included during retrieval. This highlights the critical role of external knowledge in enhancing model performance and underscores the significance of our proposed dataset.

\noindent
\textbf{Part 2: In-depth Analysis of Model and Prompt Performance:}
Having established the need for our approach of prompting with the right context, we now analyze the performance of various models and prompting strategies on the NCERT-QA dataset with the full context provided. As shown in the Table \ref{tab:bestperf}, there is a clear correlation between model size and performance. The larger models, Llama4-Scout-17B and DeepSeek-70B, significantly outperform their smaller, open-source counterparts. Between the two models, Llama4-Scout-17B shows higher performance, and among the small open-sourced counterparts, Llama3.2-3B leads largely in F1 and faithfulness. 

The choice of prompting strategy also had a profound impact. The meta oneshot prompt emerged as the most consistently effective strategy, delivering the best F1 scores for most models (See Table \ref{tab:bestperf}) in both English and Science subsets. This suggests that a high-quality, machine-generated instruction combined with a single, clear example offers an optimal balance of guidance. Conversely, the Chain-of-Thought strategy yielded surprisingly poor results, particularly on the Faithfulness metric across models(e.g., 0.532 for Llama4-Scout; See Appendix \ref{sec:appendix}). This indicates that encouraging detailed step-by-step reasoning for this task caused the models to "hallucinate" details beyond the provided context. Comparing performance across subjects, models consistently scored slightly higher on the Science dataset than the English dataset. For instance, Llama4-Scout achieved a faithfulness score of 0.87 on Science, while its performance on English was closer to 0.85. This suggests that the factual, descriptive nature of the science texts is more amenable to prompting with context than the inferential and narrative complexities of the English literary texts.

\begin{table}[htbp]
\centering
\caption{Comparison of Small (S) and Large (L) models: (a) Performance without contextual data; (b) Best model and prompt type on English and Science with context.}
\tiny
\begin{minipage}[t]{0.48\textwidth}
\centering
\subcaption{}\label{tab:nocon}
\begin{tabular}{clcccc}
\toprule
\textbf{NCERT} & \textbf{Model} & \textbf{F1} & \textbf{B-F1} & \textbf{R-L} & \textbf{SS} \\
\midrule
\multirow{5}{*}{\rotatebox[origin=c]{90}{English}}
& Llama4 (L) & 0.32 & 0.81 & 0.27 & 0.78 \\
& DS (L) & 0.13 & 0.78 & 0.07 & 0.77 \\
& Gemma3 (S) & 0.26 & 0.79 & 0.22 & 0.77 \\
& Nemo (S) & 0.24 & 0.79 & 0.19 & 0.77 \\
& Llama3.2 (S) & 0.18 & 0.76 & 0.14 & 0.70 \\
\midrule
\multirow{5}{*}{\rotatebox[origin=c]{90}{Science}}
& Llama4 (L) & 0.47 & 0.81 & 0.41 & 0.88 \\
& DS (L) & 0.46 & 0.88 & 0.40 & 0.89 \\
& Gemma3 (S) & 0.27 & 0.81 & 0.23 & 0.79 \\
& Nemo (S) & 0.25 & 0.89 & 0.20 & 0.79 \\
& Llama3.2 (S) & 0.19 & 0.78 & 0.15 & 0.72 \\
\bottomrule
\end{tabular}
\end{minipage}%
\hfill
\begin{minipage}[t]{0.48\textwidth}
\centering
\subcaption{}\label{tab:bestperf}
\begin{tabular}{cllccccc}
\toprule
\textbf{NCERT} & \textbf{Model} & \textbf{Prompt} & \textbf{F1} & \textbf{B-F1} & \textbf{R-L} & \textbf{SS} & \textbf{FF} \\
\midrule
\multirow{5}{*}{\rotatebox[origin=c]{90}{English}} 
& Llama4 (L) & MP-1S & \textbf{0.46} & 0.86 & 0.40 & 0.86 & \textbf{0.85} \\
& DS (L) & MP & 0.45 & 0.86 & 0.39 & 0.87 & 0.79 \\
& Llama3.2 (S) & MP-1S & \textbf{0.40} & 0.84 & 0.35 & 0.83 & \textbf{0.81} \\
& Nemo (S) & MP & 0.36 & 0.83 & 0.28 & 0.82 & 0.76 \\
& Gemma3 (S) & MP-1S & 0.35 & 0.83 & 0.31 & 0.81 & 0.80 \\
\midrule
\multirow{5}{*}{\rotatebox[origin=c]{90}{Science}} 
& Llama4 (L) & MP-1S & \textbf{0.47} & 0.88 & 0.41 & 0.88 & \textbf{0.87} \\
& DS (L) & MP-1S & 0.46 & 0.87 & 0.40 & 0.89 & 0.81 \\
& Llama3.2 (S) & MP-1S & \textbf{0.41} & 0.86 & 0.35 & 0.85 & \textbf{0.83} \\
& Nemo (S) & MP-1S & 0.37 & 0.85 & 0.29 & 0.84 & 0.78 \\
& Gemma3 (S) & MP-1S & 0.36 & 0.85 & 0.32 & 0.83 & 0.82 \\
\bottomrule
\end{tabular}
\end{minipage}
\label{tab:combined}
\end{table}

\begin{figure}
    \centering
    \includegraphics[width=1\textwidth]{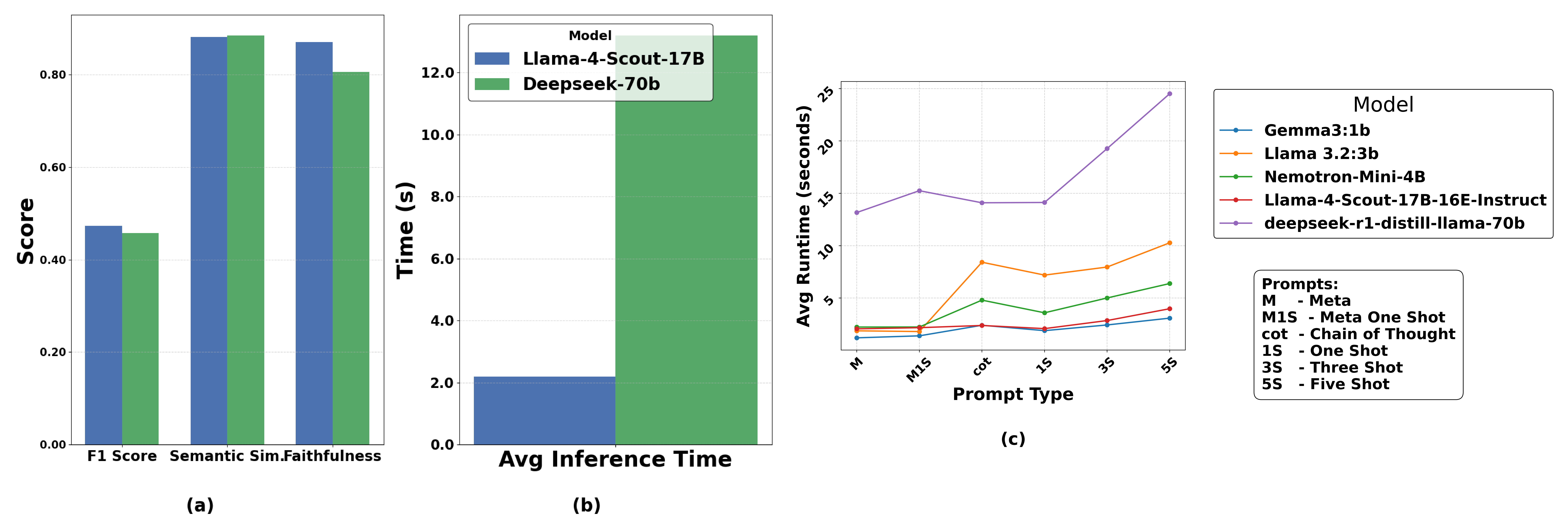}
    \caption{Performance–efficiency trade-off: (a,b) Llama-4-Scout matches DeepSeek-70B performance at far lower inference time. (c) Average runtime per prompt type shows meta and meta-one-shot with reduced latency; Llama models and Gemma3-1B are fastest, DeepSeek-70B slowest.}
    \label{fig:timetaken}
\end{figure}

\noindent
\textbf{Part 3: Practical Implications for Real-World Deployment}
While larger models deliver higher performance, a practical educational tool must also be efficient and cost-effective. In this section, we analyze the trade-off between performance and inference speed to identify the most viable model for deployment in a school setting. Fig: \ref{fig:timetaken} (a) and (b) compares our two top-performing models. While DeepSeek-70B holds a slight edge in semantic similarity, Llama4-Scout-17B achieves a better F1 score and is significantly more faithful, all while being over 6 times faster. An average inference time of ~2 seconds is well within the acceptable range for a real-time interactive assistant, whereas a ~13 second wait is likely too slow for an engaging student experience.
This analysis demonstrates that Llama4-Scout-17B is not just a compromise but arguably the superior choice for this application. It delivers state-of-the-art results on the metrics that matter most (accuracy and faithfulness) with the efficiency required for practical, cost-effective deployment in schools, where computational resources may be limited. Additionally as can be seen from Fig. \ref{fig:timetaken} (c), among the smaller open-source models, Gemma3-1B and Llama3.2-3B, exhibit lower latency, with Llama3.2-3B offering a more optimal trade-off between latency and overall performance. 

\section{Conclusion}
\label{sec:conc}
In this study, we introduce the NCERT-QA dataset to bridge the gap between curriculum content and educational Q\&A systems. Our categorization of question types highlights the diversity present in the dataset, ensuring a comprehensive representation of curriculum-based queries. This work marks an initial step toward expanding the dataset across additional grades and subjects. Through extensive evaluations across multiple models and prompting strategies, we emphasize the vital role of high-quality, curriculum-aligned data in enhancing the accuracy and relevance of responses. Our findings demonstrate that incorporating contextual knowledge significantly improves model performance, reinforcing the importance of structured retrieval and carefully curated datasets such as NCERT-QA. Our analysis of models of varying scales demonstrates the potential of smaller open-source models for practical deployment in resource-constrained environments. Our dataset and pipeline, PustakAI, establishes a foundation for building more robust and scalable educational AI systems that are closely aligned with academic curriculum. More detailed observations and exhaustive analysis will be provided on the ArXiv version of this paper.

\begin{credits}
\subsubsection{\ackname} We express our gratitude towards Bebras Challenge for making the practice question available in the public domain for practice. We acknowledge BITS Pilani's OPERA and New Faculty Seed Grant and the CSIS department's infrastructural support. ACM for bringing this challenge to India and promoting it.  

\end{credits}

\appendix
\section{Appendix: Comparison of Model Performance Under Contextual and Non-Contextual Prompts}
\label{sec:appendix}
Comparative performance of models visualized across all prompting strategies.
\begin{figure}[H]
    \centering
    \includegraphics[width=0.8\textwidth]{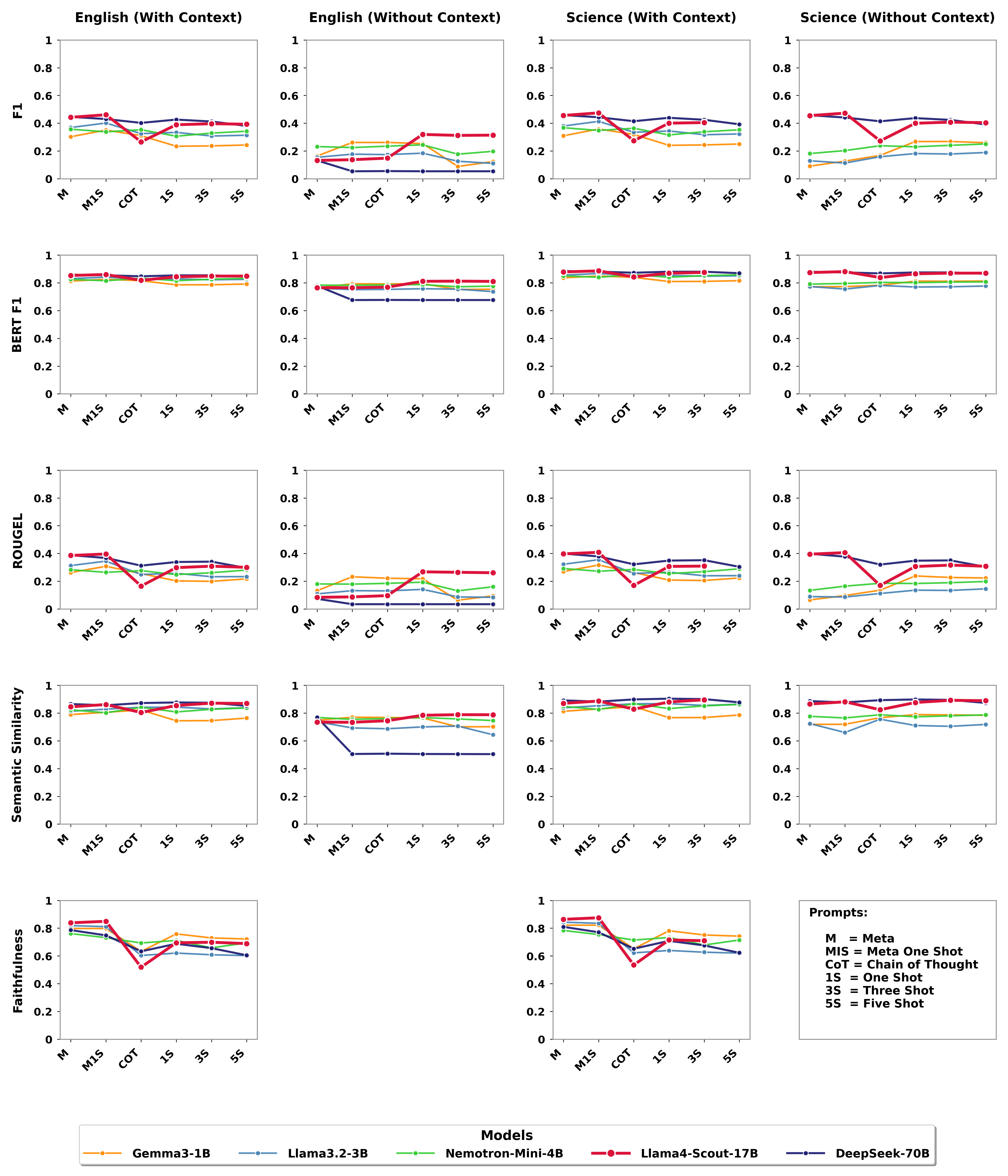}
    \caption{Performance of models on contextual vs. non-contextual prompts. Faithfulness metric is absent for non-contextual prompts since faithfulness measures the match between the provided context and the LLM’s generated answer, which cannot be computed without context.}
    \label{fig:final_graph}
\end{figure}

\bibliographystyle{splncs04}
\bibliography{mybibliography}

\end{document}